\def\year{2020}\relax
\documentclass[letterpaper]{article} % DO NOT CHANGE THIS
\usepackage{aaai20}  % DO NOT CHANGE THIS
\usepackage{times}  % DO NOT CHANGE THIS
\usepackage{helvet} % DO NOT CHANGE THIS
\usepackage{courier}  % DO NOT CHANGE THIS
\usepackage[hyphens]{url}  % DO NOT CHANGE THIS
\usepackage{graphicx} % DO NOT CHANGE THIS
\usepackage{graphicx}  % DO NOT CHANGE THIS
\usepackage{amsfonts}
\usepackage{amsmath}
\newcommand{\tabincell}[2]{\begin{tabular}{@{}#1@{}}#2\end{tabular}}
\usepackage{multirow}
\usepackage{booktabs}
\usepackage{tabu}
\usepackage{tabularx}
\title{Pointwise Rotation-Invariant Network with\protect\\ Adaptive Sampling and 3D Spherical Voxel Convolution}
\author{Yang You,\textsuperscript{\rm 1}\thanks{Contributed equally to this paper.} Yujing Lou,\textsuperscript{\rm 1}\footnotemark[1] Qi Liu,\textsuperscript{\rm 1}\\ \Large{\textbf{Yu-Wing Tai,\textsuperscript{\rm 2} Lizhuang Ma,\textsuperscript{\rm 1} Cewu Lu,\textsuperscript{\rm 1} Weiming Wang\textsuperscript{\rm 1}\thanks{Weiming Wang is the corresponding author.}}} \\ % All authors must be in the same font size and format. Use \Large and \textbf to achieve this result when breaking a line
\textsuperscript{\rm 1}Shanghai Jiao Tong University, China\\ %If you have multiple authors and multiple affiliations
\textsuperscript{\rm 2}Tencent, China\\
\{qq456cvb, louyujing, enerald, wangweiming, ma-lz, lucewu\}@sjtu.edu.cn, yuwingtai@tencent.com % email address must be in roman text type, not monospace or sans serif
}
 
\begin{document}

\maketitle

\begin{abstract}
Point cloud analysis without pose priors is very challenging in real applications, as the orientations of point clouds are often unknown. In this paper, we propose a brand new point-set learning framework PRIN, namely, \textbf{P}ointwise \textbf{R}otation-\textbf{I}nvariant \textbf{N}etwork, focusing on rotation-invariant feature extraction in point clouds analysis. We construct spherical signals by Density Aware Adaptive Sampling to deal with distorted point distributions in spherical space. In addition, we propose Spherical Voxel Convolution and Point Re-sampling to extract rotation-invariant features for each point. Our network can be applied to tasks ranging from object classification, part segmentation, to 3D feature matching and label alignment. We show that, on the dataset with randomly rotated point clouds, PRIN demonstrates better performance than state-of-the-art methods without any data augmentation. We also provide theoretical analysis for the rotation-invariance achieved by our methods. 

\end{abstract}
\section{Introduction}
Deep learning on point clouds has received tremendous interest in recent years. Since depth cameras capture point clouds directly, efficient and robust point processing methods like classification, segmentation and reconstruction have become key components in real-world applications. Robots, autonomous cars, 3D face recognition and many other fields rely on learning and analysis of point clouds. 

\begin{figure}[h]
\begin{center}
  \includegraphics[width=\columnwidth]{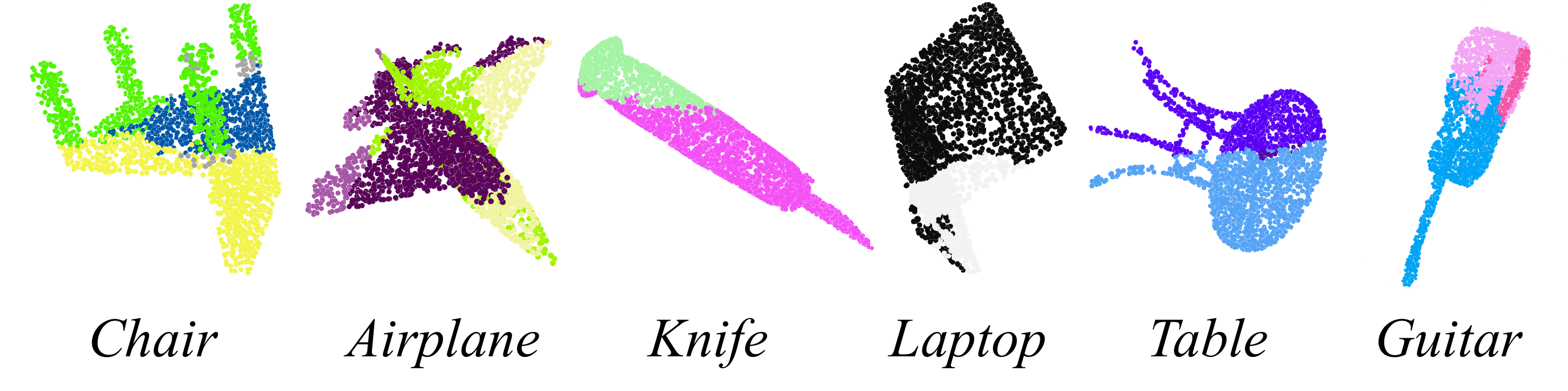}
\end{center}
\caption{\textbf{PointNet++ part segmentation results on rotated shapes.} When trained on objects with canonical orientations and evaluated on rotated ones, PointNet++ is unaware of their orientations and fails to segment their parts out.}
\label{fig:pn_partition}
\end{figure}

Existing works like PointNet~\cite{8099499} and PointNet++~\cite{qi2017pointnet++} have achieved remarkable results in point cloud learning and shape analysis. But they focus on objects with canonical orientations. In real applications, these methods fail to be applied to rotated shape analysis since the model orientation is often unknown as a priori, as shown in Figure~\ref{fig:pn_partition}. In addition, existing frameworks require massive data augmentation to handle rotations, which induces unacceptable computational cost.

Spherical CNN~\cite{cohen2018spherical} and a similar method~\cite{esteves2018learning} try to solve this problem and propose a global feature extracted from continuous meshes, while they are not suitable for point clouds since they project 3D meshes onto their enclosing spheres using a ray casting scheme. Difficulty lies in how to apply spherical convolution in continuous domain to sparse point clouds. Besides, by projecting onto unit sphere, their method is limited to processing convex shapes, ignoring any concave structures. Therefore, we propose a pointwise rotation-invariant network (PRIN) to handle these problems. Firstly, we observe the discrepancy between unit spherical space and Euclidean space, and propose Density Aware Adaptive Sampling (DAAS) to avoid biases. Secondly, we come up with Spherical Voxel Convolution (SVC) without loss of rotation-invariance, which is able to capture any concave information. Furthermore, we propose Point Re-sampling module that helps to extract rotation-invariant features for each point. 

PRIN is a network that directly takes point clouds with random rotations as the input, and predicts both categories and pointwise segmentation labels without data augmentation. It absorbs the advantages of both Spherical CNN and PointNet-like network by keeping rotation-invariant features, while maintaining a one-to-one point correspondence between the input and output. 
PRIN learns rotation-invariant features at spherical voxel grids. Afterwards, these features could be aggregated into a global descriptor or per-point descriptor to achieve model classification or part segmentation, respectively.

We experimentally compare PRIN with various state-of-the-art approaches on the benchmark dataset: ShapeNet part dataset~\cite{Yi16} and ModelNet40~\cite{wu20153d}. Additionally, PRIN can be applied to 3D point matching and label alignment. PRIN exhibits remarkable performance on all these tasks.

\begin{figure*}[t]
\begin{center}
\includegraphics[width=\textwidth]{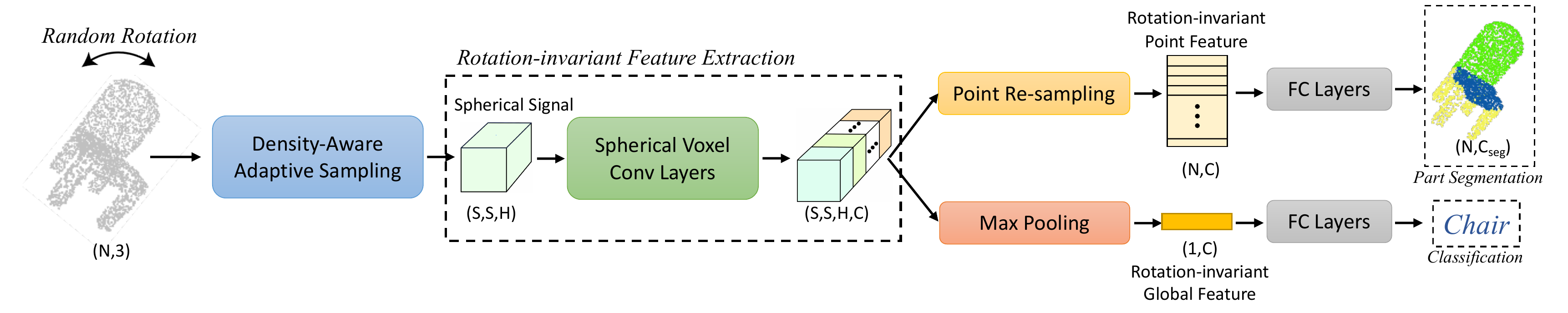}
\end{center}
%TODO: caption
\caption{\textbf{PRIN Architecture.}  
Our network takes sparse points as input, and then uses Density-Aware Adaptive Sampling to transform the signals into spherical voxel grids. This spherical voxel signals is then passed through several Spherical Voxel Convolution layers, ending with a feature at each spherical voxel grid. Any point feature can be extracted by point re-sampling, which is used to do pointwise part segmentation. All these voxel features can also be maxpooled to get a global feature, which is suitable for classification.
}
\label{fig:pipeline}
\end{figure*}

The key contributions of this paper are as follows:
\begin{itemize}
\item{We design a novel deep network processing pipeline that extracts rotation-invariant point-level features.}
\item{Three key modules: Density Aware  Adaptive  Sampling (DAAS), Spherical Voxel Convolution (SVC) and Point Re-sampling are proposed. }
\item{We show that our network can be used for point cloud part segmentation, classifcation, 3D point matching and label alignment under different rotations.}
The code to reproduce our results is available online.\footnote{https://github.com/qq456cvb/PRIN}
\end{itemize}

%------------------------------------------------------------------------

\section{Related Work}

\subsection{Rotation-invariant Features}
Rotation-Invariance is often regarded as a preliminary to the success of template matching and object detection, in both 2D and 3D domains.

The development of rotation-invariant features from geometries could be retrospected to manual designed features, including Structural Indexing~\cite{stein1992structural}, Signature of Histogram Orientations (SHOT)~\cite{shot} and CGF~\cite{khoury2017learning}. They construct a local reference frame (LRF) which aligns the model into its canonical pose in order to extract rotation-invariant point features. However, these methods depend on local surface variations, therefore are sensitive to noise and point densities. Besides, these descriptors rely on delicate hand-craft design, and could only capture low-level geometric features. For a more complete review on traditional feature descriptors, we refer to \cite{guo20143d}.

Recently, some papers consider generalizations of 2D CNNs that exploit larger groups of symmetries~\cite{gens2014deep,cohen2016steerable}, including the 2D/3D rotation group~\cite{dieleman2015rotation}. Spherical CNN~\cite{cohen2018spherical} and a similar method~\cite{esteves2018learning} propose to extract global rotation-invariant features from continuous meshes, while they are not suitable for point clouds since they project 3D meshes onto enclosing spheres using a ray casting scheme.

\subsection{Deep Learning on 3D Models}
As the consequence of success in deep learning, various methods have been proposed for better understanding 3D models. Convolutional neural networks are applied to volumetric data since its format is similar to pixel and easy to transfer to existing frameworks. 3D ShapeNet~\cite{wu20153d} and VoxNet~\cite{7353481} are pioneers introducing fully-connected networks to voxels. However, dealing with voxel data requires large memory and its sparsity also makes it challenging to extract particular features. 

Another research branch is multi-view methods. MVCNN~\cite{su2015multi} renders 3D models into multi-view images and propagates these images into traditional convolutional neural networks. 
These approaches are limited to simple tasks like classification and not suitable for 3D part segmentation, key point matching or other tasks. 

Dealing with point clouds directly is another popular branch, among which PointNet~\cite{8099499} is the pioneer in building a general framework for learning point clouds. Since then, many structures are proposed to learn from point clouds. PointNet++~\cite{qi2017pointnet++} extends PointNet by intrducing a hierarchical structure. RSNet~\cite{huang2018recurrent} is combines a novel slice pooling layer, Recurrent Neural Network (RNN) layers, and a slice unpooling layer, to better propogate features among different parts. SplatNet~\cite{su18splatnet} uses sparse bilateral convolutional layers to enable hierarchical and spatially-aware feature learning. 
DGCNN~\cite{dgcnn} proposes EdgeConv, which explicitly constructs a local graph and learns the embeddings for the edges in both Euclidean and semantic spaces.
As a follow-up, LDGCNN~\cite{zhang2019linked} links the hierarchical features from different dynamic graphs based on DGCNN.
SO-Net~\cite{li2018so} models the spatial distribution of point cloud by building a Self-Organizing Map (SOM) and its receptive field can be systematically adjusted
by conducting point-to-node k nearest neighbor search.
SpiderCNN~\cite{xu2018spidercnn} designs the convolutional filter as a product of a simple step function that captures local geodesic information and a Taylor polynomial that ensures
the expressiveness.
DeepSets~\cite{zaheer2017deep} provides a family of functions which are permutation-invariant and gives a competitive result on point clouds.
Point2Sequence~\cite{liu2019point2sequence} uses a recurrent neural network (RNN)
based encoder-decoder structure, where an attention mechanism is proposed to highlight the importance of different area scales.
Kd-Network~\cite{8237361} utilizes kd-tree structures to form the computational graph, which learns from point clouds hierarchically. However, few of them are robust to random rotations, making it hard to apply them to real applications.

%------------------------------------------------------------------------
\section{Method}
\subsection{Problem Statement}
Given a set of unordered points $\mathcal{X} = \{x_i\}$ with $x_i \in \mathbb{R}^{d^{in}}$ and $i=1,\cdots,N$, where $N$ denotes the number of input points and $d^{in}$ denotes the dimensionality of input features at each point, which can be positions, colors, etc. Our goal is to produce a set of point-wise features $\mathcal{Y} = \{y_i\}$ with $y_i \in \mathbb{R}^{d^{out}}$ and $i=1,\cdots,N$, which are invariant to input orientations. PRIN can be modeled as a rotation-invariant function $\mathcal{F}:\mathcal{X}\mapsto\mathcal{Y}$. Although the problem and the method developed are general, we focus on the case $d_{in}=3$ using only Euclidean coordinates as the input. To implement function $\mathcal{F}$, we design mainly three modules: (1) Density Aware Adaptive Sampling (DAAS) module $\Gamma: \mathbb{R}^{N\times 3}\rightarrow\mathbb{R}^{S^2\times H}$ that constructs spherical signals; (2) Spherical Voxel Convolution (SVC) module $\Phi: \mathbb{R}^{S^2\times H\times C_{in}}\rightarrow\mathbb{R}^{S^2\times H\times C_{out}}$ that extracts rotation-invariant features; (3) Point Re-sampling module $\Lambda: \mathbb{R}^{S^2\times H\times C_{out}}\rightarrow\mathbb{R}^{N\times C_{out}}$ that re-samples points from spherical signals. We will explain these modules in the following sections and the whole pipeline is shown in Figure~\ref{fig:pipeline}.

\subsection{Density Aware Adaptive Sampling}
\label{sec:input}
In this step, the objective is to build spherical signals from irregular point clouds. Nonetheless, if we sample point clouds uniformly into regular spherical voxels, we will meet a problem: points around the pole appear to be more sparse than those around the equator in spherical coordinates, which brings a bias to the resulting spherical voxel signals. 
To address this problem, we propose Density Aware Adaptive Sampling (DAAS). DAAS leverages a non-uniform filter to adjust the density discrepancy brought by spherical coordinates, thus reducing the bias. This process can be modeled as $\Gamma:\mathbb{R}^{N\times 3}\rightarrow\mathbb{R}^{S^2\times H}$.

\paragraph{Unit Spherical Space $S^2\times H$.} A point in unit spherical space is identified by $(\alpha, \beta, h)$, where $(\alpha, \beta) \in S^2$ denotes the polar angle and the azimuthal angle while $h\in H$ represents the radial distance to the sphere center.

\begin{figure}[h]
\begin{center}
    \includegraphics[width=\columnwidth]{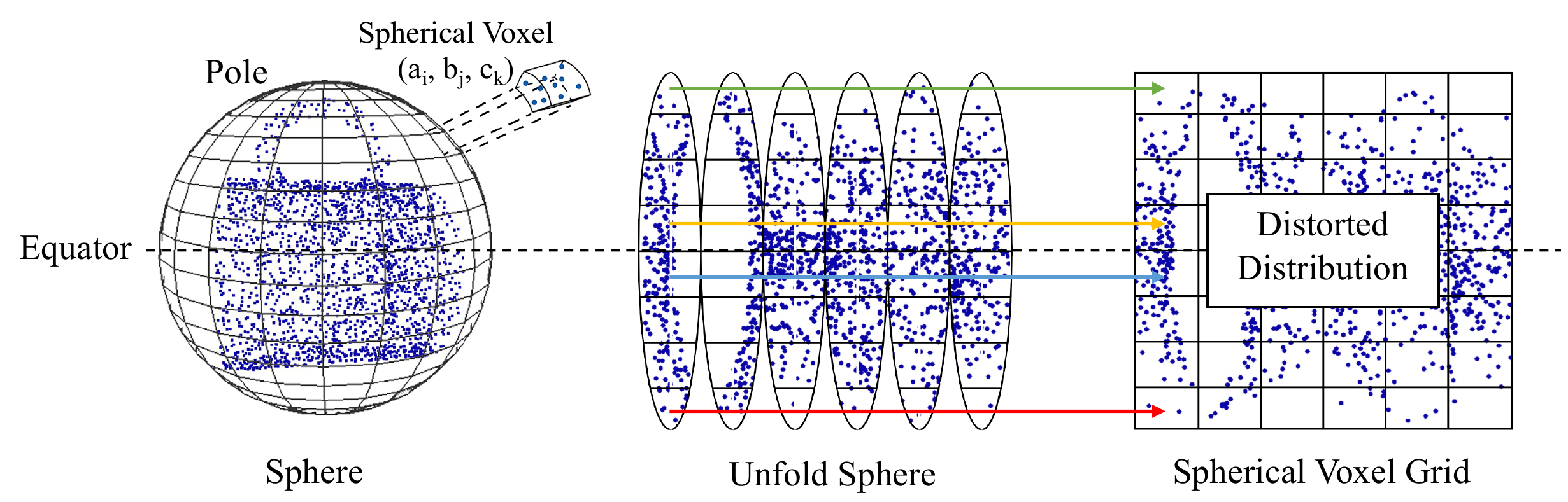}
\end{center}
    \caption{Distorted distribution in spherical coordinates.}
\label{fig:density}
\end{figure}

In practice, we divide unit spherical space into spherical voxels, which are indexed by $(i, j, k) \in I\times J\times K$, where $I\times J\times K$ is the spatial resolution, also known as the bandwidth~\cite{kostelec2007soft}. Each spherical voxel is represented by its center $(a_i, b_j, c_k)$, where $(a_i$, $b_j)$ is spherical coordinate with $a_i\in [0,2\pi]$, $b_j\in[0,\pi]$, and $c_k\in[0,1]$ is the distance to the center of unit sphere. The division of spherical space is shown in the left of Figure \ref{fig:density}.

We first map each point $x_i\in \mathcal{X}$ from Euclidean space $\mathbb{R}^3$ to unit spherical space $S^2\times H$ by calculating its spherical coordinates $(\alpha_i,\beta_i,h_i)$,
and then calculate the spherical signal $f : S^2\times H \rightarrow\mathbb{R}$ as follows:
\begin{equation}
\label{eq:change}
\begin{split}
    f(a_i, b_j, c_k) = & \frac{\sum\limits^N_{n=1}w_n\cdot(\delta - \|h_n - c_k\|)}{\sum\limits^N_{n=1}w_n},
\end{split}
\end{equation}
where $w_n$ is a normalizing factor that is defined as
\begin{equation}
\label{eq:wt}
\begin{split}
    w_n =\  &\mathbf{1}(\|\alpha_n - a_i\| < \delta) \\ 
            \cdot &\mathbf{1}(\|\beta_n - b_j\| < \eta\delta) \\ 
            \cdot &\mathbf{1}(\|h_n - c_k\| < \delta),
\end{split}
\end{equation}
$\delta$ is a predefined threshold filter width and $\eta$ is a density aware factor explained in the next subsection.
We use $(\delta - \|h_n - c_k\|) \in [0, \delta]$  in Equation~\ref{eq:change} because it captures information along $H$ axis, which is orthogonal to $S^2$, making it invariant under random rotations. 

\paragraph{Density Aware Factor.} $\eta = sin(\beta)$ is the density aware sampling factor. We treat spherical voxel signals as a tensor $T \in \mathbb{R}^{I\times J\times K}$ with equal-sized grids. However, these grids are not equal-sized in Euclidean space, making any point distributions distorted in spherical voxel grids, as shown in Figure~\ref{fig:density}. For a more rigorous proof of the factor $sin(\beta)$, please see our supplementary material.

\subsection{Spherical Voxel Convolution} 
\label{sec:rotinv}
Given some spherical voxel signal $f: S^2\times H\rightarrow\mathbb{R}^{C_{in}}$, we introduce Spherical Voxel Convolution (SVC) to further refine the features, while keeping it rotation-invariant. This step implements the operator $\Phi: \mathbb{R}^{S^2\times H \times C_{in}} \rightarrow \mathbb{R}^{S^2\times H \times C_{out}}$.

Compared with Spherical CNN~\cite{esteves2018learning} where only spherical signals defined in $S^2$ get convoluted. We propose the spherical voxel convolution which takes spherical signals defined in $S^2\times H$ as the input. 
To understand spherical voxel convolution, we first give some definitions and annotations. Here, we only consider functions with $C_{in}=C_{out}=1$ to reduce clutter, while extensions to higher-dimension functions are straght-forward.

\paragraph{Rotations.} The 3D rotation group, often denoted $SO(3)$, which is termed "special orthogonal group", is the group of all rotations about the origin of three-dimensional Euclidean space $\mathbb{R}^3$ under the operation of composition. SO(3) can be parameterized by ZYZ-Euler angles $(\alpha, \beta, \gamma)$, where $\alpha \in [0, 2\pi], \beta \in [0, \pi], $ and $\gamma \in [0, 2\pi]$. For $R\in SO(3)$ and any function $g : SO(3)\rightarrow\mathbb{R}$, according to \cite{kyatkin2000engineering}, we have
\begin{equation}
\label{eq:so3invariance}
    \int_{SO(3)}g(Rx)dx=\int_{SO(3)}g(x)dx.
\end{equation}

To process some spherical signal $g : S^2\times H \rightarrow\mathbb{R}$, we establish a bijective homorphism (isomorphism) between $S^2\times H$ and $SO(3)$ by considering $H$ as $SO(3)/S^2=SO(2)$ (see supplementary for more details). Similarly, we have
\begin{equation}
\label{eq:s2hinvariance}
    \int_{S^2\times H}g(Rs)ds=\int_{S^2\times H}g(s)ds.
\end{equation}

\paragraph{Rotations of Spherical Voxel Signals.} In order to implement the rotation-invariant feature extraction, we introduce a rotation operator $L_R$ on the signal $g : S^2\times H\rightarrow \mathbb{R}$,
\begin{equation}
\label{eq:rotvox}
    [L_Rg](s) = g(R^{-1}s),
\end{equation}
where $s\in S^2\times H$. 

\paragraph{Spherical Voxel Convolution.} Now we give a formal definition of spherical voxel convolution, evaluated at point $p\in S^2\times H$:
\begin{equation}
    \label{eq:voxelconv}
    \begin{split}
        [\psi\star f] (p) = &\langle L_{P}\psi, f\rangle \\
        = &\int_{S^2\times H}\psi(P^{-1}s)f(s)ds,
    \end{split}
\end{equation}
where $f$ denotes the input signals, $\psi$ is the filter and $P$ is the corresponding element of $p$ in $SO(3)$ (by the isomorphism between $S^2\times H$ and $SO(3)$ defined above).

\paragraph{Rotation Invariance.}
Suppose that the input point cloud is rotated by an arbitrary rotation $R$, we have $\forall p\in S^2\times H$, $p\rightarrow Rp$. As a consequence, we have the corresponding spherical signal rotated: $f\rightarrow L_Rf$,  since $f$ is sampled from the input point cloud. We are now ready to prove the rotation-invariance by applying SVC to the rotated input:
\begin{equation}
    \label{eq:rotinv}
    \begin{split}
        [\psi\star L_Rf] (Rp) = &\langle L_{RP}\psi, L_Rf\rangle \\
        = &\int_{S^2\times H}\psi(P^{-1}R^{-1}s)f(R^{-1}s)ds\\
        = &\int_{S^2\times H}\psi(P^{-1}s)f(s)ds\\
        = &[\psi\star f] (p),
    \end{split}
\end{equation}
where the second last step follows from Eq.~\ref{eq:s2hinvariance}. It is obvious that SVC extracts the same feature no matter how the input point cloud rotates, which ensures the \textbf{rotation-invariance}.

In practice, with analogy to $SO(3)$ convolution, Spherical Voxel Convolution (SVC) can be efficiently computed by Fast Fourier Transform (FFT)~\cite{kostelec2008ffts}. Convolutions are implemented by first doing FFT to convert both the input and filters into spectral domain, then multiplying them and converting results back to spatial domain, using Inverse Fast Fourier Transform (IFFT)~\cite{kostelec2008ffts}.

\subsection{Point Re-sampling}
\label{sec:output}
After Spherical Voxel Convolution (SVC), we re-sample features  at the location of original points, with \textit{Trilinear Interpolation} as our operator $\Lambda : \mathbb{R}^{S^2\times H \times C}\rightarrow\mathbb{R}^{N \times C}$. Each point's feature is a weighted average of nearest eight voxels, where the weights are inversely related to the distances to these spherical voxels. Then they are passed through fully connected layers to get the pointwise features.  

\subsection{Architecture}
To summarize, our rotation-invariant function $\mathcal{F}$ is a concatenation of the three modules/operators mentioned above: $\mathcal{Y} = \Lambda(\Phi(\Gamma(\mathcal{X})))$. We first transform points from Euclidean space to unit spherical space by operator $\Gamma$, and then conduct a rotation-invariant feature transform by operator $\Phi$, and finally re-sample the point features in Euclidean space by operator $\Lambda$.

After extracting pointwise rotation-invariant features, we are able to do part segmentation by concatenating some fully connected layers. Our network could also realize object classification by placing a different head. In this case, we maxpool all the features in spherical voxels and pass this global feature through several fully connected layers, as shown in Figure~\ref{fig:pipeline}.

\begin{table*}[t]
\begin{center}
\resizebox{\textwidth}{!}{
\begin{tabular}{l|c|c|cccccccccccccccc|c|c}
\toprule[1pt]
\multirow{2}*{}  & \multicolumn{18}{c|}{Arbitrary Rotation} & \multicolumn{2}{c}{No Rotation}\\
\cline{2-21}
~ & \tabincell{c}{avg.\\ inst.} & \tabincell{c}{avg.\\ cls.}  & \tabincell{c}{air\\plane} & bag & cap & car & chair & \tabincell{c}{ear\\ phone} & guitar & knife & lamp & laptop & \tabincell{c}{motor\\ bike} & mug & pistol & rocket & \tabincell{c}{skate\\ board} & table & \tabincell{c}{avg.\\ inst.} & \tabincell{c}{avg.\\ cls.} \\
\hline
PointNet & 31.30 & 29.38 & 19.90 & 46.25 & 43.27	& 20.81	& 27.04 & 15.63 & 34.72	& 34.64	& 42.10	& 36.40	& 19.25	& 49.88	& 33.30	& 22.07	& 25.71 & 29.74 & 83.15 & 78.95\\
PointNet++ & 36.66 & 35.00 & 21.90 & 51.70 & 40.06 & 23.13 & 43.03 & 9.65 & 38.51 & 40.91 & 45.56 & 41.75 & 18.18 & 53.42 & 42.19 & 28.51 & 38.92 & 36.57 & 84.63 & 81.52 \\
RS-Net & 50.38 & 32.99 & 38.29 & 15.45 & 53.78 & 33.49 & \textbf{60.83} & 31.27 & 9.50 & 43.48 & 57.37 & 9.86 & 20.37 & 25.74 & 20.63 & 11.51 & 30.14 & 66.11 & 84.92 & 81.41\\
PCNN & 28.80 & 31.72  & 23.46 & 46.55 & 35.25 & 22.62 & 24.27 & 16.67 & 32.89 & 39.80 & 52.18 & 38.60 & 18.54 & 48.90 & 27.83 & 27.46 & 27.60 & 24.88  & 85.13 & 81.80\\
SPLATNet & 32.21 & 38.25 & 34.58 & 68.10 & 46.96 & 19.36 & 16.25 & 24.72 & \textbf{88.39} & 52.99 & 49.21 & 31.83 & 17.06 & 48.56 & 21.20 & 34.98 & 28.99 & 28.86 & 84.97 & 82.34 \\
DGCNN & 43.79 & 30.87 & 24.84 & 51.29 & 36.69 & 20.33 & 30.07 & 27.86 & 38.00 & 45.50 & 42.29 & 34.84 & 20.51 & 48.74 & 26.25 & 26.88 & 26.95 & 28.85 & 85.15 & 82.33 \\
SO-Net & 26.21 & 14.37 & 21.08 & 8.46 & 1.87 & 11.78 & 27.81 & 11.99 & 8.34 & 15.01 & 43.98 & 1.81 & 7.05 & 8.78 & 4.41 & 6.38 & 16.10 & 34.98 & 84.83 & 81.16 \\
SpiderCNN & 31.81 & 35.46 & 22.28 & 53.07 & 54.2 & 22.57 & 28.86 & 23.17 & 35.85 & 42.72 & 44.09 & \textbf{55.44} & 19.23 & 48.93 & 28.65 & 25.61 & 31.36 & 31.32 & \textbf{85.33} & \textbf{82.40} \\
\hline
SHOT+PointNet & 32.88 & 31.46 & 37.42 & 47.30 & 49.53 & 27.71 & 28.09 & 16.34 & 9.79 & 27.66 & 37.33 & 25.22 & 16.31 & 50.91 & 25.07 & 21.29 & 43.10 & 40.27 & 32.75 & 31.25 \\
CGF+PointNet & 50.13 & 46.26 & 50.97 &\textbf{70.34} & 60.44 & 25.51 & 59.08 & 33.29 & 50.92 & 71.64 & 40.77 & 31.91 & 23.93 & 63.17 & 27.73 & 30.99 & 47.25 & 52.06 & 50.13 & 46.31 \\
\hline
\textbf{Ours} & \textbf{58.76} & \textbf{54.22} & \textbf{51.02} & 56.18 & \textbf{63.39} & \textbf{40.78} & 57.31 & \textbf{54.39} & 58.07 & \textbf{71.67} & \textbf{61.82} & 38.63 & \textbf{24.93} & \textbf{70.44} & \textbf{45.89} & \textbf{48.68} & \textbf{58.73} & \textbf{65.59} & 71.52 & 70.33 \\
\bottomrule[1pt]
\end{tabular}}
\end{center}   
\caption{Shape part segmentation test results mIoU (\%) on the randomly rotated ShapeNet part dataset.}
\label{tab:compare_miou}
\end{table*}

%-------------------------------------------------------------------------
\section{Experiments}
\label{seq:experiments}

In this section, we arrange comprehensive expreriments to evaluate PRIN for point cloud analysis on different tasks. First, we demonstrate that our model can be applied to 3D shape part segmentation and classification with random rotations. Then, we provide some applications on 3D point matching and shape alignment. At last, we conduct ablation study to validate the design of each part.

\paragraph{Implementation Details.} PRIN is implemented in Python using PyTorch on one NVIDIA TITAN Xp. 
In all of our experiments, Adam optimization algorithm is employed for training, with a batch size of 16. The learning rate begins with 0.01 and decays with a rate of 0.5 every 5 epochs. During training we set the input bandwidth on $S^2$ to 32 and resolution on $H$ to 64. The threshold filter width $\delta=1/32$.

\subsection{Part Segmentation on Rotated Shapes}
\label{seq:part}

\paragraph{Dataset.} ShapeNet part dataset~\cite{Yi16} contains 16,881 shapes from 16 categories in which each shape is annotated with expert verified part labels from 50 different labels in total. Most shapes are composed of two to five parts. We follow the data split in PointNet~\cite{8099499}.

Part segmentation is a challenging task for rotated shape analysis. We train each model in Table \ref{tab:compare_miou} on the non-rotated training set and evaluate them on the non-rotated and rotated test set. We show that our pipeline can accomplish part segmentation task on the randomly rotated test set without seeing these rotations. Even though state-of-the-art deep learning networks, like PointNet++~\cite{qi2017pointnet++}, DGCNN~\cite{dgcnn}, SpiderCNN~\cite{xu2018spidercnn}, etc., can achieve fairly good results, they all fail on the rotated point clouds. Besides, traditional local features based on local reference frames (LRF) like CGF~\cite{khoury2017learning} and SHOT~\cite{shot} are also compared. We train a PointNet-like network with these local features as the input. Although these descriptors are rotation-invariant, their performance is inferior to ours.

\begin{figure}[t]
\begin{center}
\includegraphics[width=\columnwidth]{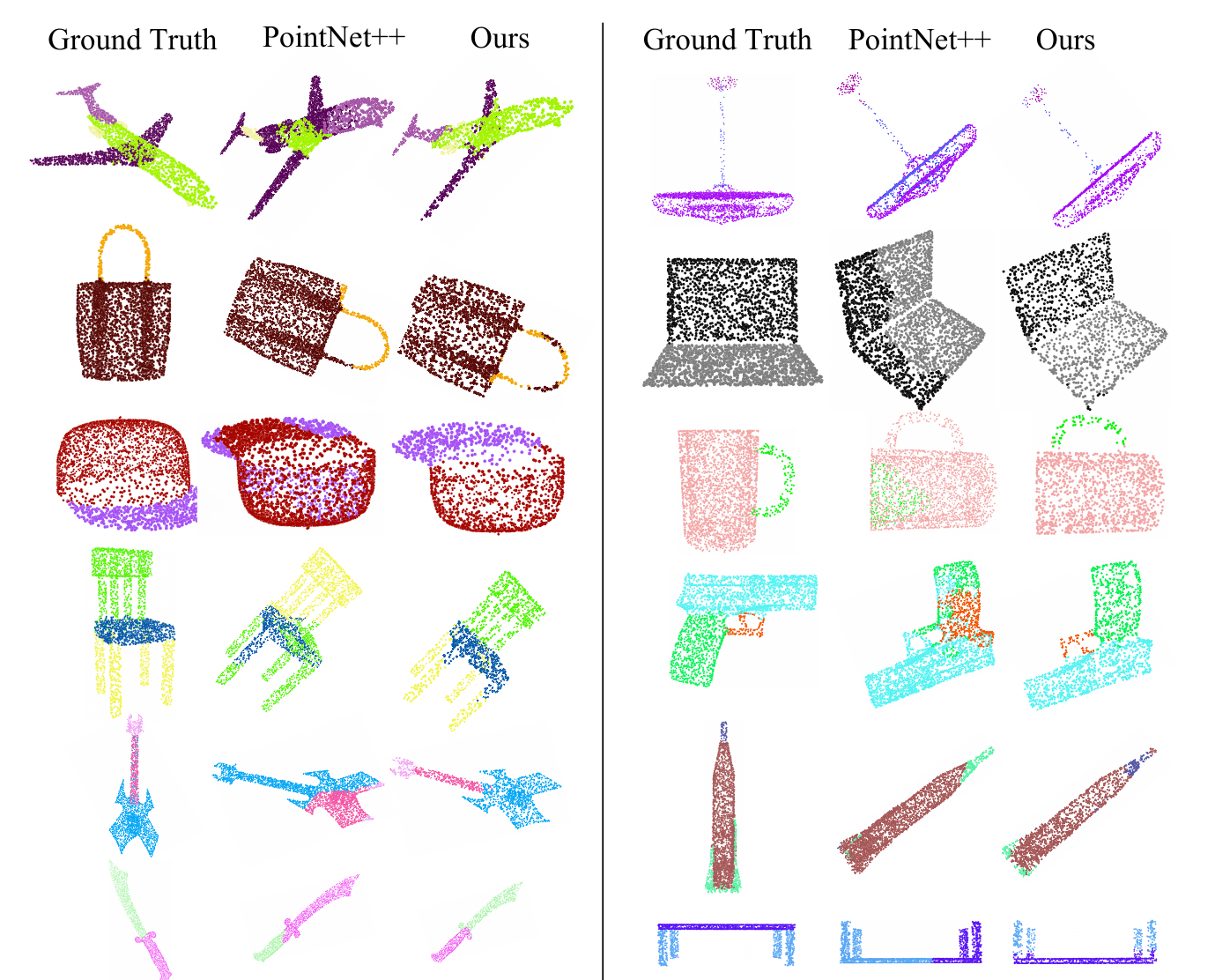}
\end{center}
\caption{\textbf{Visualization of results.} We test PointNet++ and PRIN on rotated point clouds trained on the canonical orientation dataset. Our network generalizes well on unseen orientations.}
\label{fig:main}
\end{figure}

Table~\ref{tab:compare_miou} shows the results of PRIN and the compared networks. All the results are illustrated in the metric of mean IoU. We can find that for those compared methods, mIoU decreases drastically when tested on the rotated point clouds. 
In addition, PRIN shows the better performance than rotation-invariant local features like CGF and SHOT. Figure~\ref{fig:main} gives the visualization of test results between state-of-the-art deep learning methods and PRIN over the ShapeNet part dataset. Influenced by the canonical orientation of point clouds in the training set, networks like PointNet++ just learn a simple partition of Euclidean space, regardless of how objects are positioned in their real pose.

There exists a performance gap of our method when evaluated on the non-rotated and rotated test set, shown in the first two and last two columns of Table \ref{tab:compare_miou}. This  is  because,  when  we  are  implementing  spherical voxel convolution, the continuous domain is discretized up to a limited resolution.  However, when evaluated on the rotated test set, rotations do not necessarily lie on these centers of voxels, resulting in a small gap.

In this task, we use four SVC layers with output channels 64, 40, 40, 50 in our experiments. All convolution layers have the same bandwidth 32. Each kernel $\psi$ has non-local support, where $\psi(\alpha, \beta, h)$ iff $\beta=\pi/2$ and $h = 0$. After point re-sampling, two fully-connected layers of size 50 and 50 are connected at the end. The final network contains $\approx$ 0.4M parameters and takes 12 hours to train, for 40 epochs.

%-------------------------------------------------------------------------

\subsection{Classification on Rotated Shapes}
\paragraph{Dataset.} ModelNet40~\cite{wu20153d}, 3D shape classification dataset, contains 12,308 shapes from 40 categories. Here, we use its corresponding point clouds dataset provided by PointNet~\cite{8099499}.

\begin{table}[h]
\begin{center}
\begin{tabularx}{\columnwidth}{p{0.3\columnwidth}|p{0.15\columnwidth}<{\centering}|p{0.15\columnwidth}<{\centering}|p{0.15\columnwidth}<{\centering}|p{0.15\columnwidth}<{\centering}}
\toprule[1pt]
Method & Input & NR & AR  & Params \\
\hline
PointNet & \multirow{9}*{\tabincell{c}{point\\xyz}} & 88.45 & 12.47 & 3.5M \\
PointNet++ & ~ & 89.82 & 21.35 & 1.5M \\
Point2Sequence & ~ & 92.60 & 10.53 & 1.8M \\
Kd-Network & ~ & 86.20 & 8.49 & 3.6M \\
Spherical CNN  & ~ & 81.73 & 55.62 & 0.4M\\
DeepSets & ~ & 88.73 & 9.72 & 0.8M \\
LDGCNN & ~ & 92.91 & 17.82 & 2.8M \\
SO-Net & ~ & \textbf{93.44} & 9.64 & 2.7M \\
\textbf{Ours} & ~ &  80.13 & \textbf{70.35} & \textbf{0.4M} \\
\hline
SHOT+PointNet & \multirow{2}*{\tabincell{c}{local\\ features}}& 48.79 & 48.79 & 1.1M \\
CGF+PointNet & ~ & 57.70 & 57.89 & 4.2M \\
\bottomrule[1pt]
\end{tabularx}
\end{center}
\caption{\textbf{Classification results on ModelNet40 dataset.} Performance is evaluated in  accuracy. NR means to train with no rotations and test with no rotations. AR means to train with no rotations and test with arbitrary rotations. PRIN is robust to arbitrary rotations while other methods fail to classify correctly.}

\label{tab:classify}
\end{table}
Though classification does not require pointwise rotation-invariant features but a global feature, our network still benefits from DAAS and SVC, so that PRIN could handle point clouds with unknown orientations. 

We compare our network with several state-of-the-art deep learning methods that take point Euclidean coordinates and local rotation-invariant features as the input. We train our network on the non-rotated training set and achieve 70.35\% accuracy on the rotated test set. Almost all other deep learning methods fail to generalize to unseen orientations. Besides, local rotation-invariant features like CGF and SHOT are inferior to our method. The results are shown in Table~\ref{tab:classify}.

For this task, we use four Spherical Voxel Convolution (SVC) layers with channels 64, 50, 70, 350 in our experiments. The bandwidths for each layer are 64, 32, 22, 7. Each kernel $\psi$ has non-local support, where $\psi(\alpha, \beta, h)$ iff $\beta=\pi/2$ and $h = 0$. A maxpooling layer is connected after SVC layers to get a global feature, followed by two fully-connected layers. The final network contains $\approx$ 0.4M parameters and takes 12 hours to train, for 40 epochs.

\subsection{Application}

\begin{figure}[h]
\begin{center}
  \includegraphics[width=\columnwidth]{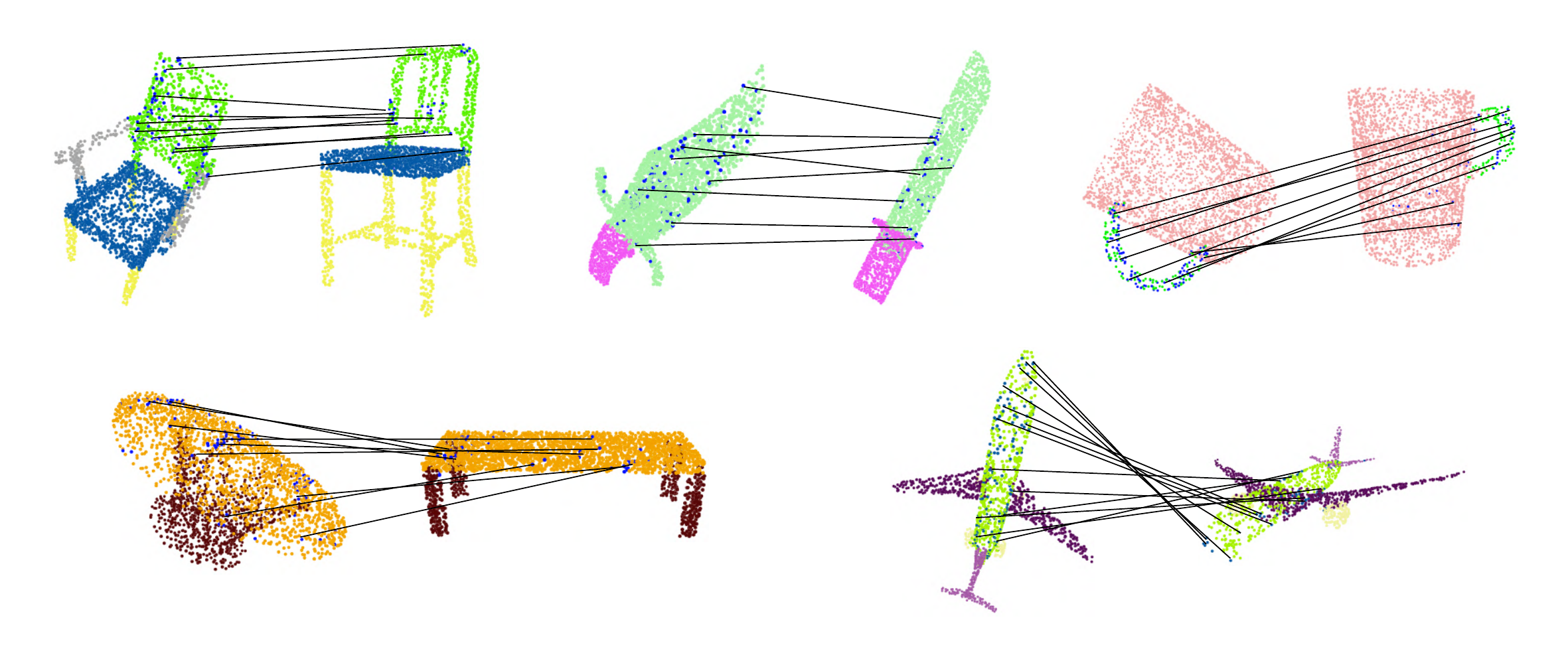}
\end{center}
\caption{\textbf{3D point matching.} Point matching results between the query object features (left) and the retrieved ones (right) under different orientations.}
\label{fig:matching}
\end{figure}

\noindent\textbf{3D Rotation-invariant Point Descriptors.}
For 3D objects, SHOT, CGF and other local features exploits local reference frames (LRF) as local descriptors. Similarly, PRIN is able to produce high quality rotation-invariant 3D point descriptors, which is pretty useful as pairwise matching become possible regardless of rotations. In this application, we can retrieve the closest descriptor from similar objects under arbitrary orientations. After that, we can perform point matching according to rotation-invariant point features. The demonstration is shown in Figure \ref{fig:matching}. 
We know where each point of the query object locates on another object. Moreover, such 3D point descriptors could have the potentiality to do scene searching and parsing as the degree of freedom reduces from six to three, leaving only translations.

To test the matching accuracy on ShapeNet part dataset, we compute and store pointwise features of 80\% test objects (non-rotated) in ShapeNet as a database. Then we compute pointwise features of the other 20\% test objects that are randomly rotated as queries. Feature retrievals are done by finding queries' nearest neighbours in the database. We evaluate three different methods to see if points corresponding to the same part are matched. The results are summarized in Table~\ref{tab:desc}.
\begin{table}[h]
\begin{center}
\begin{tabular}{l|c}
\hline
Method & Matching Accuracy \\
\hline
PointNet & 38.91\\
SHOT & 17.11 \\
CGF & 52.51\\
\hline
\textbf{Ours} & \textbf{64.23} \\
\hline
\end{tabular}
\end{center}
\caption{\textbf{3D descriptor matching results.} Accuracy is the number of matched points corresponding to the same part divided by the number of all matched points.}
\label{tab:desc}
\end{table}

\begin{figure}[h]
\begin{center}
  \includegraphics[width=0.9\columnwidth]{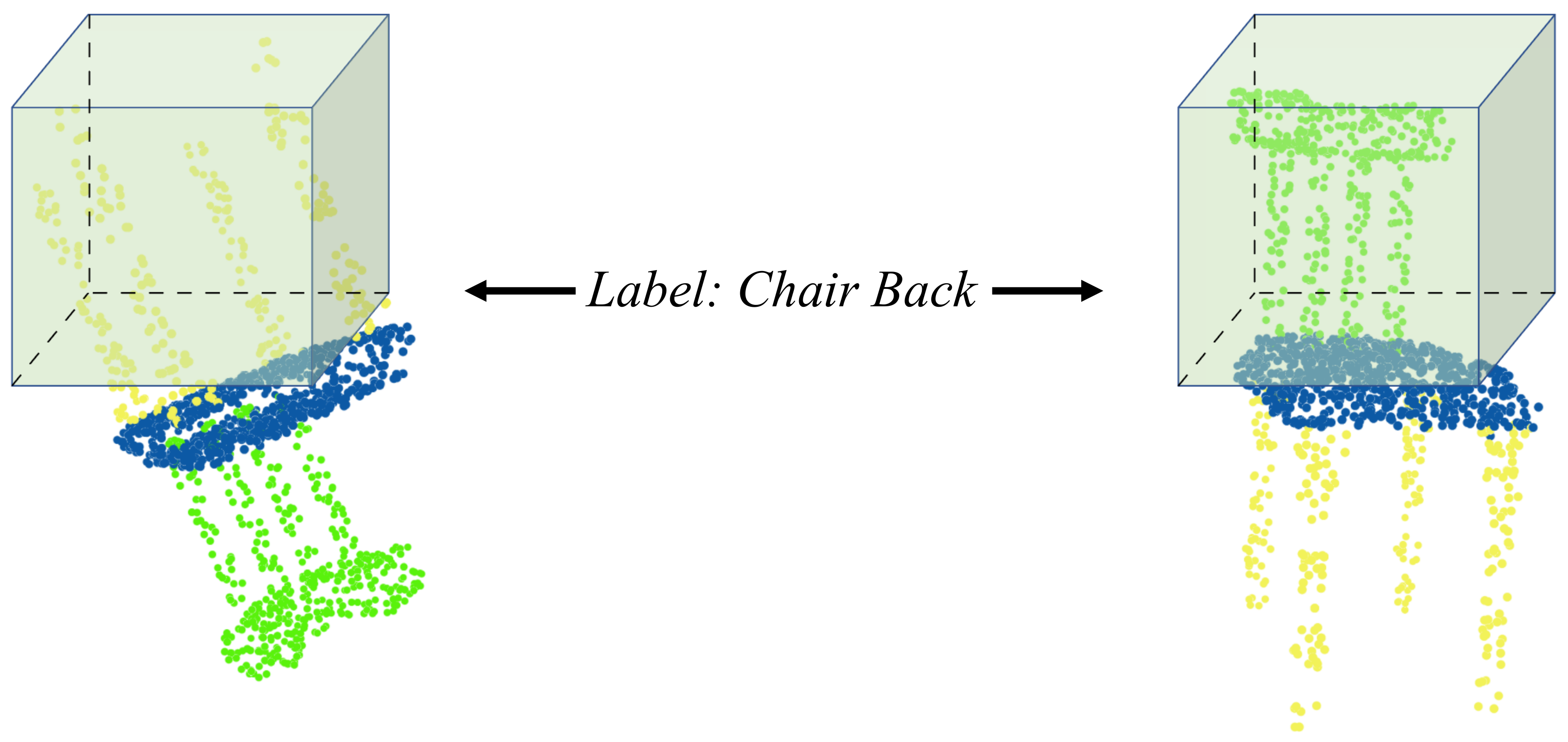}
\end{center}
\caption{\textbf{Chair alignment with its back on the top.} \textbf{Left:} A misalignment induces large KL divergence (1.34). \textbf{Right:} Required labels fulfilled with small KL divergence (0.12).}
\label{fig:alignment}
\end{figure}

\noindent\textbf{Shape Alignment with Label Priors.}
We now introduce a task that given some label requirements in the space, our network would align the point cloud to satisfy these requirements. For example, one may want a chair that has its back on the top. So we add the virtual points describing the label requirement. Once the KL divergence between predicted scores and ground-truth one-hot labels of these virtual points is minimized, the chair is aligned with its back on the top. This is shown in Figure \ref{fig:alignment}.
%------------------------------------------------------------------------

\subsection{Ablation Study}

\begin{figure}[h]
\begin{center}
   \includegraphics[width=0.95\columnwidth]{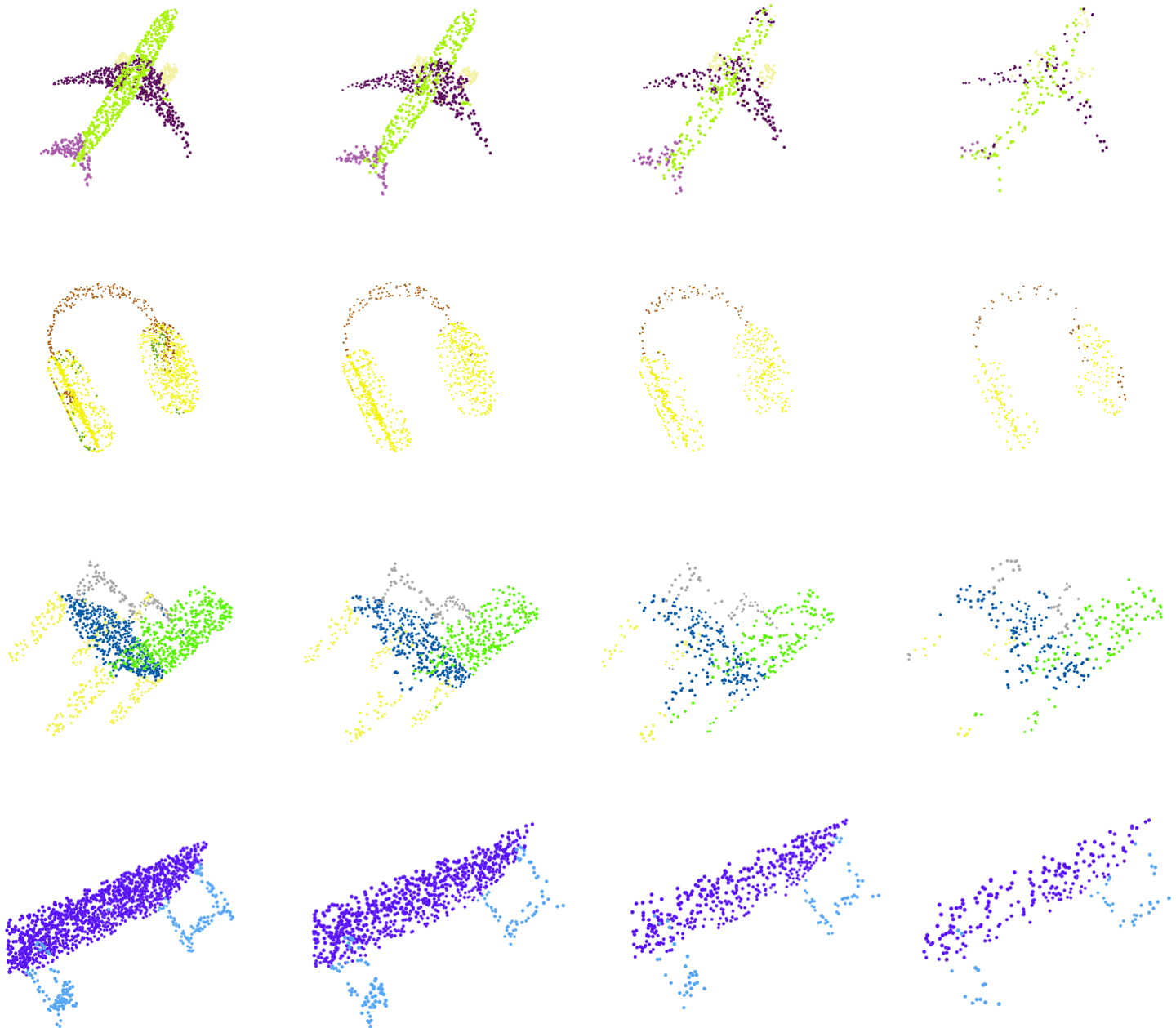}
\end{center}
   \caption{\textbf{Segmentation robustness results.} \textbf{From left to right}: we sample a subset of 2048, 1024, 512, 256 points from test point clouds respectively. We observe that our network is robust to missing points and gives consistent results.}
\label{fig:robustness}
\end{figure}

In this section we evaluate numerous variations of our method to verify the rationality of network design. Experiment results are summarized in Table \ref{tab:ablation} and Figure \ref{fig:robustness}.
\begin{table}[h]
\begin{center}
\begin{tabularx}{\columnwidth}{p{0.2\columnwidth}<{\centering}|p{0.2\columnwidth}<{\centering}|p{0.2\columnwidth}<{\centering}|p{0.3\columnwidth}<{\centering}}
\toprule[1pt]
\tabincell{c}{bandwidth\\ on $S^2$} & \tabincell{c}{resolution\\ on $H$} & DAAS & acc/mIoU \\
\hline
\textbf{32} & \textbf{64} & \textbf{Yes} & \textbf{79.43/58.76} \\
16 & 64 & Yes & 74.53/53.80 \\
8 & 64 & Yes & 71.17/47.10 \\
\hline
32 & 32 & Yes & 76.56/55.63 \\
32 & 1 & Yes & 76.19/54.32 \\
\hline
32 & 64 & No & 74.61/54.23 \\
\bottomrule[1pt]
\end{tabularx}
\end{center}
\caption{\textbf{Ablation study.} Test accuracy of PRIN on rotated ShapeNet part dataset. The model with various bandwidths, resolutions on $H$ and whether to use DAAS are tested.}
\label{tab:ablation}
\end{table}

\paragraph{Input Bandwidth.} One decisive factor of our network is the bandwidth. Bandwidth is used to describe the sphere precision, which is also the resolution on $S^2$. Mostly, large bandwidth offers more details of spherical voxels, such that our network can extract more specific point features of point clouds. While large bandwidth assures more specific representation of part knowledge, more memory cost is accompanied. Increasing bandwidth from 8 to 32 leads to a relative improvement of 11.61\% and 24.76\% on accuracy and mIoU, which is shown in Table \ref{tab:ablation}.

\paragraph{Resolution on $H$.} Here we study the effects of the resolution on $H$ dimension, which is also the number of sphere layers that are stacked. Table~\ref{tab:ablation} shows the results of different numbers of resolutions we set. We find that increasing the resolution from 1 to 64 improves the accuracy and mIoU results by 4.25\% and 8.17\% relatively.

\paragraph{Sampling Strategy.} Recall that in Equation~\ref{eq:change}, we construct our signals on each spherical voxel with an density aware sampling filter. We now study the effect of Density Aware Adaptive Sampling (DAAS) and the results are shown in Table \ref{tab:ablation}. We see that using the $sin(\beta)$ corrected sampling filter gives superior performance, which confirms our theory.

\paragraph{Segmentation Robustness.}
PRIN also demonstrates an excellent robustness to corrupted and missing points. Although the density of point clouds declines, our network still segments correctly for each point. In Figure \ref{fig:robustness}, PRIN predicts consistent labels regardless of the point density. 
%-------------------------------------------------------------------------

\section{Discussions and Future Work}

\begin{figure}[h]
\begin{center}
  \includegraphics[width=\columnwidth]{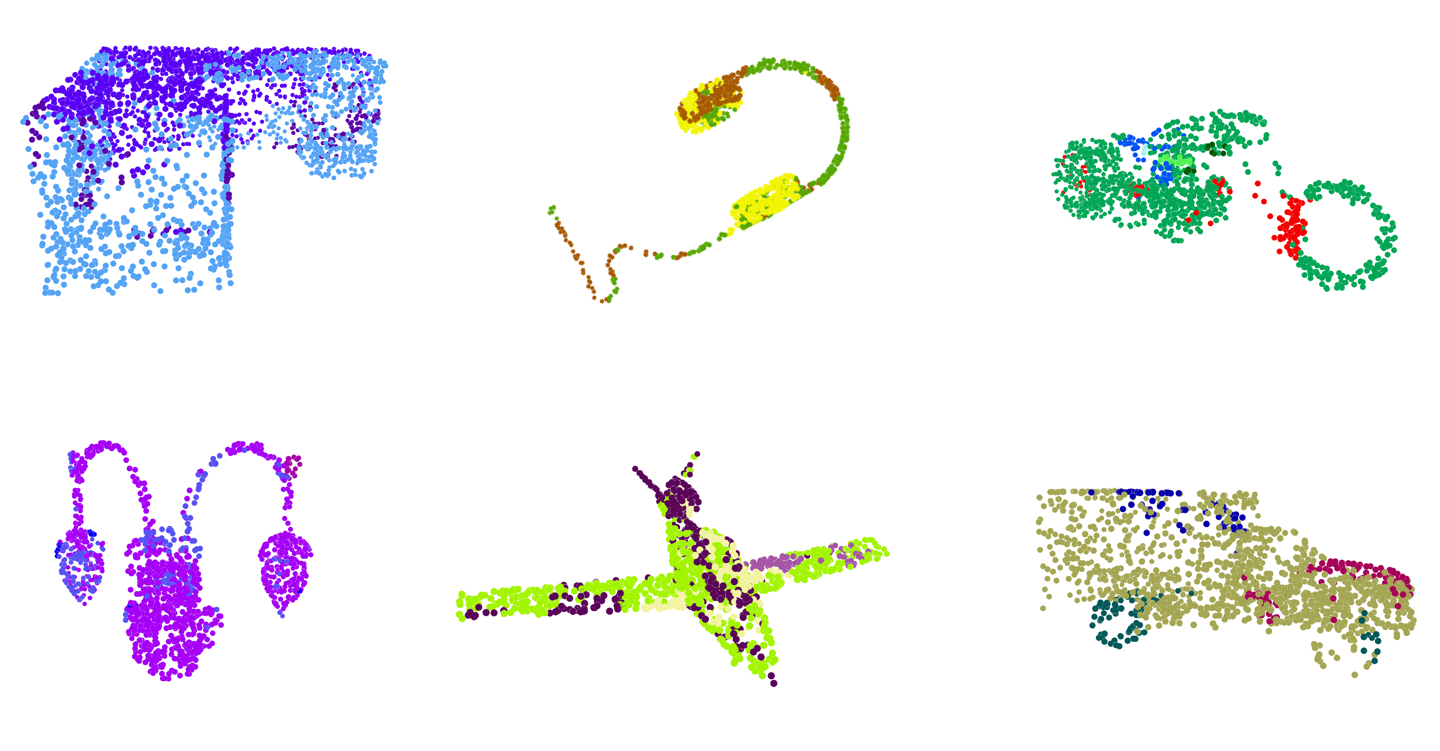}
\end{center}
\caption{\textbf{Failure cases.}}
\label{fig:failure}
\end{figure}

Though our network is invariant to point cloud rotations, we see there are some failure cases that when there are complex internal structures of the object as in Figure~\ref{fig:failure}. This may be caused by that our filters are not perfect and special filters can be designed. Despite current filters are density aware, they are not aware of curvature changes. Also, due to computational considerations, the input voxel resolution, which is defined by the bandwidth~\cite{cohen2018spherical}, is limited to about 32 while better results can be obtained with higher resolutions. We leave this memory-efficient convolution and special design of filters as our future work.

%------------------------------------------------------------------------

\section{Conclusion}

We present PRIN, a network that takes any input point cloud and leverages Density Aware Adaptive Sampling (DAAS) to construct signals on spherical voxels. Then Spherical Voxel Convolution (SVC) and Point Re-sampling follow to extract pointwise rotation-invariant features. We place two different output heads to do both 3D point clouds classification and part segmentation. Our experiments show that our network is robust to arbitrary orientations even not seeing them. Our network can be applied to 3D point feature matching and shape alignment with label priors. We show that our model can naturally handle arbitrary input orientations for different tasks and provide theoretical analysis to help understand our network.
\section{Acknowledgements}
This work is supported in part by the National Key R\&D Program of China (No. 2017YFA0700800), and the National Natural Science Foundation of China (No. 61772332, 51675342, 51975350, 61972157).
\appendix
\section*{Supplementary}

\section{Density-Aware Factor $\eta$}
\label{sec:freqchange}
\paragraph{Spacing Representations}
We denote volumes (spacing) in euclidean ($\mathbb{R}^3$) and spherical ($S^2$) coordinates as $dxdydz$ and $d\alpha d\beta dr$ respectively, where $r=1$ is a dummy variable representing the radius. 

\paragraph{Jacobian} Given the relationship from spherical coordinates to Euclidean coordinates,
\begin{align}
    \begin{split}
        x &= rsin(\beta)cos(\alpha)\\
        y &= rsin(\beta)sin(\alpha)\\
        z &= rcos(\beta)
    \end{split}
\end{align}

The Jacobian $J_t = \frac{dxdydz}{d\alpha d\beta dr}$ of this transformation is 
\begin{equation}
\begin{bmatrix}
    \frac{\partial x}{\partial\alpha}      & \frac{\partial x}{\partial\beta}  & \frac{\partial x}{\partial r} \\
    \frac{\partial y}{\partial\alpha}      & \frac{\partial y}{\partial\beta}  & \frac{\partial y}{\partial r} \\
    \frac{\partial z}{\partial\alpha}      & \frac{\partial z}{\partial\beta}  & \frac{\partial z}{\partial r}.
\end{bmatrix}
\end{equation}
Write this out,
\begin{equation}
\begin{bmatrix}
    -rsin(\beta)sin(\alpha)     & rcos(\beta)cos(\alpha)  & sin(\beta)cos(\alpha) \\
    rsin(\beta)cos(\alpha)      & rcos(\beta)sin(\alpha)  & sin(\beta)sin(\alpha) \\
    0      & -rsin(\beta)  & cos(\beta)
\end{bmatrix}.
\end{equation}
The absolute value of the Jacobian determinant is $r^2sin(\beta)$. 

\paragraph{Spacing Relations}
The spacing relationship between $\mathbb{R}^3$ and $S^2$ is,
\begin{equation}
    dxdydz = r^2sin(\beta)d\alpha d\beta dr.
\end{equation}
Since $r = 1$, we have,
\begin{equation}
    dxdydz = sin(\beta)d\alpha d\beta.
\end{equation}
Therefore, we choose density-aware factor $\eta$ to be $sin(\beta)$ as density is reciprocal to spacing.

% Since the sample frequency is the inverse of spacing, we get
% \begin{equation}
%     \omega(x, y, z) = \frac{\omega(\alpha, \beta, h)}{h^2sin(\beta)}
% \end{equation}
\section{Haar Measure and Parameterization on $S^2$ and $SO(3)$}
\paragraph{Parameterization of $SO(3)$}
For any element $R \in SO(3)$, it could be parameterized by ZYZ Euler angles,
\begin{equation}
\label{eq:zyz}
    R = R(\alpha, \beta, \gamma) = Z(\alpha)Y(\beta)Z(\gamma)
\end{equation}
where $\alpha \in [0, 2\pi], \beta \in [0, \pi], $ and $\gamma \in [0, 2\pi]$, and Z/Y are rotations around Z/Y axes.

\paragraph{Haar Measure of $SO(3)$}
The normalized Haar measure is 
\begin{equation}
    dR = \frac{d\alpha}{2\pi}\frac{d\beta sin(\beta)}{2}\frac{d\gamma}{2\pi}.
\end{equation}
The Haar measure \cite{kyatkin2000engineering,nachbin1976haar} is invariant because it has the property that 
\begin{equation}
    \int_{SO(3)}f(R^{'}R)dR = \int_{SO(3)}f(R)dR,
\end{equation}
for any $R^{'}\in SO(3)$.
\paragraph{Parameterization of $S^2$}
Likewise, an element $x \in S^2$ is written as
\begin{equation}
\label{eq:zy}
    x(\alpha, \beta) = Z(\alpha)Y(\beta)n,
\end{equation}
where $n$ is the north pole.

This parameterization makes explicit the fact that the sphere is a quotient $S^2 = SO(3)/SO(2)$, where $SO(2)$ is the subgroup of rotations around the Z axes. 

\paragraph{Haar Measure of $S^2$ and $SO(2)$}
The normalized Haar measure for the sphere is 
\begin{equation}
    dx = \frac{d\alpha}{2\pi}\frac{d\beta sin(\beta)}{2}.
\end{equation}

The normalized Haar measure for $SO(2)$ is 
\begin{equation}
\label{eq:so2}
    dh = \frac{d\gamma}{2\pi}.
\end{equation}

\section{Mapping between $S^2 \times H$ and $SO(3)$}
\label{sec:map}
\paragraph{Bijactive Mapping}
For an element $(x, h) \in S^2\times H$, where $x:=x(\alpha, \beta) \in S^2$, if we view $H$ as $SO(2)$,
\begin{equation}
    (x(\alpha, \beta), h) = (Z(\alpha)Y(\beta)n, Z(h)).
\end{equation}
There is a bijective mapping from $(x, h)$ to $R(\alpha, \beta, h)$, as $R(\alpha, \beta, h)$ can be written as,
\begin{equation}
    R(\alpha, \beta, h) = Z(\alpha)Y(\beta)Z(h),
\end{equation}
and the mapping:
\begin{equation}
    Z(\alpha)Y(\beta)Z(h) \Longleftrightarrow (Z(\alpha)Y(\beta)n, Z(h)).
\end{equation}
\paragraph{Isomorphism by Rotation Operator}
With this mapping, any rotation that happens in voxel space $S^2\times H$ will transfer to $SO(3)$ safely,
\begin{equation}
\begin{split}
    (Qx(\alpha, \beta), h) &=(QZ(\alpha)Y(\beta)n, Z(h))\\
    &\Rightarrow (QZ(\alpha)Y(\beta))Z(h)\\
    &=QZ(\alpha)Y(\beta)Z(h) \\
    &=QR(\alpha, \beta, h)
\end{split}
\end{equation}
Notice that there is a $2\pi$ constant factor change between the measure of $H$ and the measure of rotations around $Z$ axes, as shown in eq. \ref{eq:so2}.

% \clearpage
% \bigskip
\bibliographystyle{aaai} \bibliography{main.bib}

\end{document}